\documentclass[runningheads]{llncs}
\usepackage[T1]{fontenc}
\usepackage{graphicx}
\usepackage{tcolorbox}
\usepackage{relsize}
\usepackage{xspace}
\usepackage{subfiles}
\usepackage{booktabs}
\usepackage{subfiles}
\usepackage{graphicx}
\usepackage{multicol}
\usepackage{amsmath}
\usepackage{amsfonts}
\usepackage{booktabs,makecell,multirow,tabularx}
\usepackage{subfig}
\usepackage{csquotes}
\usepackage[hidelinks]{hyperref}
\usepackage{float}
\usepackage{placeins}

\begin{document}
\definecolor{mycolor}{rgb}{0,0,0}
\title{Enhancing Learned Image Compression via Cross Window-based Attention}
%
%

\author{Priyanka Mudgal \and
Feng Liu}
\authorrunning{Mudgal et al.}
%
\institute{Portland State University, Portland OR 97124, USA \\
\email{\{pmudgal,fliu\}@cs.pdx.edu}}

%

\maketitle              
%

%
%
%
\begin{abstract}
In recent years, learned image compression methods have demonstrated superior rate-distortion performance compared to traditional image compression methods. Recent methods utilize convolutional neural networks (CNN), variational autoencoders (VAE), invertible neural networks (INN), and transformers. Despite their significant contributions, a main drawback of these models is their poor performance in capturing local redundancy. Therefore, to leverage global features along with local redundancy, we propose a CNN-based solution integrated with a feature encoding module. The feature encoding module encodes important features before feeding them to the CNN and then utilizes cross-scale window-based attention, which further captures local redundancy. Cross-scale window-based attention is inspired by the attention mechanism in transformers and effectively enlarges the receptive field. Both the feature encoding module and the cross-scale window-based attention module in our architecture are flexible and can be incorporated into any other network architecture. We evaluate our method on the Kodak and CLIC datasets and demonstrate that our approach is effective and on par with state-of-the-art methods. Our code is publicly available at \url{https://github.com/prmudgal/CWAM_IC_ISVC}.

\end{abstract}

\section{\textbf{Introduction}}
Image compression is an important and highly active research topic in the field of image processing \cite{theis2022lossy,Xie2021EnhancedIE,Zhou2019EndtoendOI,liu2023learned}. With the increasing use of multimedia, lossy image compression techniques play a crucial role in efficiently storing images and videos, especially with limited hardware and network resources. Over the past years, traditional lossy image compression techniques, including JPEG \cite{Wallace1991TheJS}, JPEG2000 \cite{Taubman2013JPEG2000I}, BPG \cite{bpg}, and VVC \cite{vvc}, have achieved commendable rate-distortion (RD) performance by following a multi-step process consisting of transformation, quantization, and entropy coding.

\begin{figure}
\centering
\begin{center}
\includegraphics[width=0.65\textwidth, trim = 0cm 5cm 7cm 3cm]{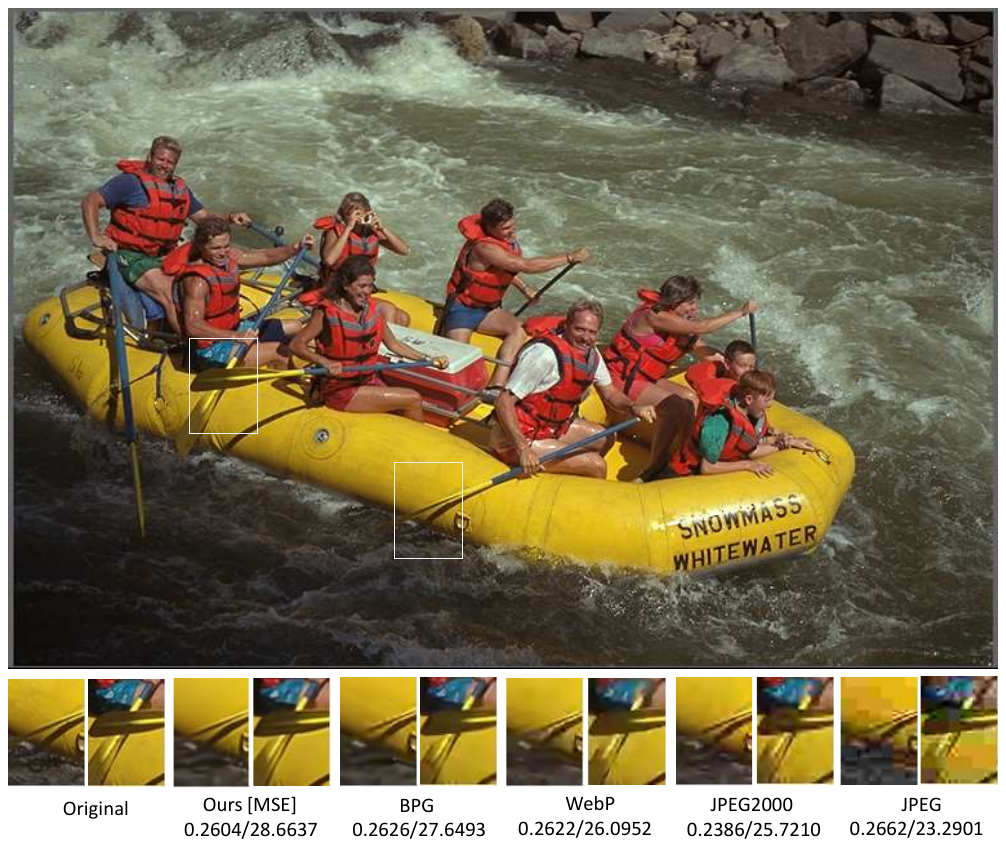}
\end{center}
\caption{\protect\label{fig:RDPlot.JPG}Visualization of decompressed images of kodim14 from Kodak dataset. It is demonstrated that our method with feature encoding module and cross-scale window-based attention is effectively compressing the image with better PSNR and optimized BPP. The subtitle shows \enquote{Method BPP$\downarrow$/PSNR$\uparrow$}.}
\end{figure}

The learned image compression (LIC) techniques \cite{kingma2022autoencoding} have been optimized end-to-end and have outperformed traditional methods based on metrics such as Peak Signal-to-Noise Ratio (PSNR) and Multi-Scale Structural Similarity (MS-SSIM) \cite{1292216}. While some recent works uses CNN-based methods with VAE \cite{Kingma2013AutoEncodingVB}, others have explored transformer-based \cite{Zou2022TheDI}, generative adversarial network (GAN) based \cite{mentzer2020highfidelity}, and INN-based \cite{Xie2021EnhancedIE} methods. All these categories of methods have achieved better RD performance \cite{Shin2006AMT,1450709} than traditional lossy image compression methods, demonstrating great opportunity for next-generation learning-based image compression techniques.

In most of the VAE-based methods, in the encoding phase, the original pixel data is converted into a lower-dimensional feature space known as the latent space. Then, it follows quantization, and later the entropy modules predict the distributions of latent variables and execute lossless coding techniques, including context-based adaptive binary arithmetic coding (CABAC) \cite{Marpe2003ContextbasedAB} or range coder (RC) \cite{Martin1979RE} to compress these variables into the bit stream. Apart from the neural network architectures used, the choice of entropy model significantly influences LIC. A range of entropy models, including hyper-prior \cite{Ball2016EndtoendOO}, auto-regressive priors \cite{Minnen2018JointAA}, and gaussian mixture model (GMM) \cite{Cheng2020LearnedIC} have evolved in recent years. These models enable entropy estimation modules to predict the distribution of latent variables, thereby enhancing rate-distortion (RD) performance.

In the decoding phase, the decoder utilizes a lossless coder such as CABAC or RC to decompress the bit stream. Subsequently, the decompressed latent variables are mapped to reconstructed images through a linear or nonlinear parametric synthesis transform. The end-to-end model combines the encoder and decoder, which can be trained together. Despite their success, these networks still face challenges related to feature distillation. CNN-based networks prioritize capturing high-level global features and sometimes struggle with learning the finer details of local features. While certain studies have addressed this issue \cite{Zou2022TheDI} by utilizing the window-based attention method, a fundamental limitation with this approach is that the window-based method uses a small receptive field, which limits the interaction between different windows and consequently limits further RD performance improvement.

Our paper addresses the aforementioned problem with existing LIC networks. First, we explore the components that provide a broader understanding of the data and a mechanism that focuses on local details. Next, we introduce a feature encoding and decoding module that improves CNNs' ability to handle complex data representations. This module includes dense blocks and convolutional layers, which strengthen feature propagation and encourage feature reuse effectively. It is integrated in a residual manner for effectiveness. Then, we adopt a modular attention module that can be combined with neural networks to capture correlations among spatially neighboring elements while considering the wider receptive field. Inspired by Cheng et al. \cite{Cheng2020LearnedIC}, Lu et al. \cite{lu2022video} and Zou et al. \cite{Zou2022TheDI}, we refer to this module as the cross window-based attention module (CWAM). This component can be integrated with CNNs to further enhance their performance. Our experiments show that the proposed method is effective and comparable to the current state-of-the-art image compression methods.

\section{\textbf{Related Work}}
For many years, traditional compression methods - namely JPEG \cite{Wallace1991TheJS}, JPEG2000 \cite{Taubman2013JPEG2000I}, WebP \cite{webp}, Better Portable Graphics (BPG) \cite{bpg}, and Versatile Video Coding (VVC) \cite{vvc} - have been widely used.  Despite their widespread use, these methods often suffer from the disadvantage of block-based compression, which results in noticeable blocking effects in reconstructed images. As these artifacts are highly visible in reconstructions produced by traditional image compression methods, learning-based image compression methods are preferred.

In recent years, learning-based image compression methods have evolved, demonstrating improved rate-distortion (RD) performance. These methods involve non-linear transformations between the image and latent feature space. Several approaches \cite{johnston2018improved,toderici2015variable,toderici2017full} based on recurrent neural networks (RNN) encode residual information from prior steps to perform image compression. However, these techniques rely on binary representation during each iteration, limiting their optimization potential in terms of bitrate. Research has also focused on VAE architectures \cite{10.5555/3294771.3294880,Ball2016EndtoendOI,Theis2017LossyIC}. These methods perform end-to-end optimization for RD performance. Subsequent studies aimed to improve entropy models for optimized rate-distortion. Balle et al. \cite{Ball2018VariationalIC} proposed a hyperprior-based entropy model that allocates additional bits to capture the distribution of latent features effectively. The hyperprior captures spatial dependencies in the latent representation by considering contextual information. Other methods \cite{Lee2018ContextadaptiveEM,8578560,NEURIPS2018_53edebc5} leverage side information to further minimize spatial redundancy in the latent space. The latest advancements include context entropy models \cite{Guo20203DCE}, channel-wise models \cite{9190935}, and hierarchical entropy models \cite{Hu_Yang_Liu_2020,NEURIPS2018_53edebc5}, which optimize the correlation of latent features. Cheng et al. \cite{9156817} introduced a gaussian mixture likelihood (GMM) to enhance accuracy, while methods incorporating CNNs with generalized divisive normalization (GDN) layers \cite{7906310} have demonstrated improved RD performance. Recent innovations integrate attention mechanisms and residual blocks into the VAE architecture \cite{9156817,Liu2019NonlocalAO,Zhang2019ResidualNA,Zhou2019EndtoendOI}, resulting in significant performance gains. Most recent techniques based on GAN \cite{Agustsson2018ExtremeLI,Rippel2017RealTimeAI,8456298}, diffusion networks \cite{theis2022lossy}, INN \cite{Xie2021EnhancedIE}, and transformers \cite{Zhou2019EndtoendOI} have shown promising results in the RD performance. Concurrent work \cite{liu2023learned} utilizes a CNN architecture combined with a transformer, proposing three methods with varying complexity, where the larger model significantly exceeds previous benchmarks in complexity.

Considering the complexity and RD performance, most of the methods perform well. However, information loss during encoding remains a persistent issue with these techniques. If information loss can be optimized, neglected information could be recovered during decoding, further enhancing RD performance. To address this, we introduce a feature encoding and decoding module that focuses more on important areas of the image and minimize information loss.

\section{\textbf{Method}}
\subsection{\textbf{Background}}
\begin{figure}
\centering
\includegraphics[width=1\textwidth, trim = 0cm 6cm 0cm 4cm]{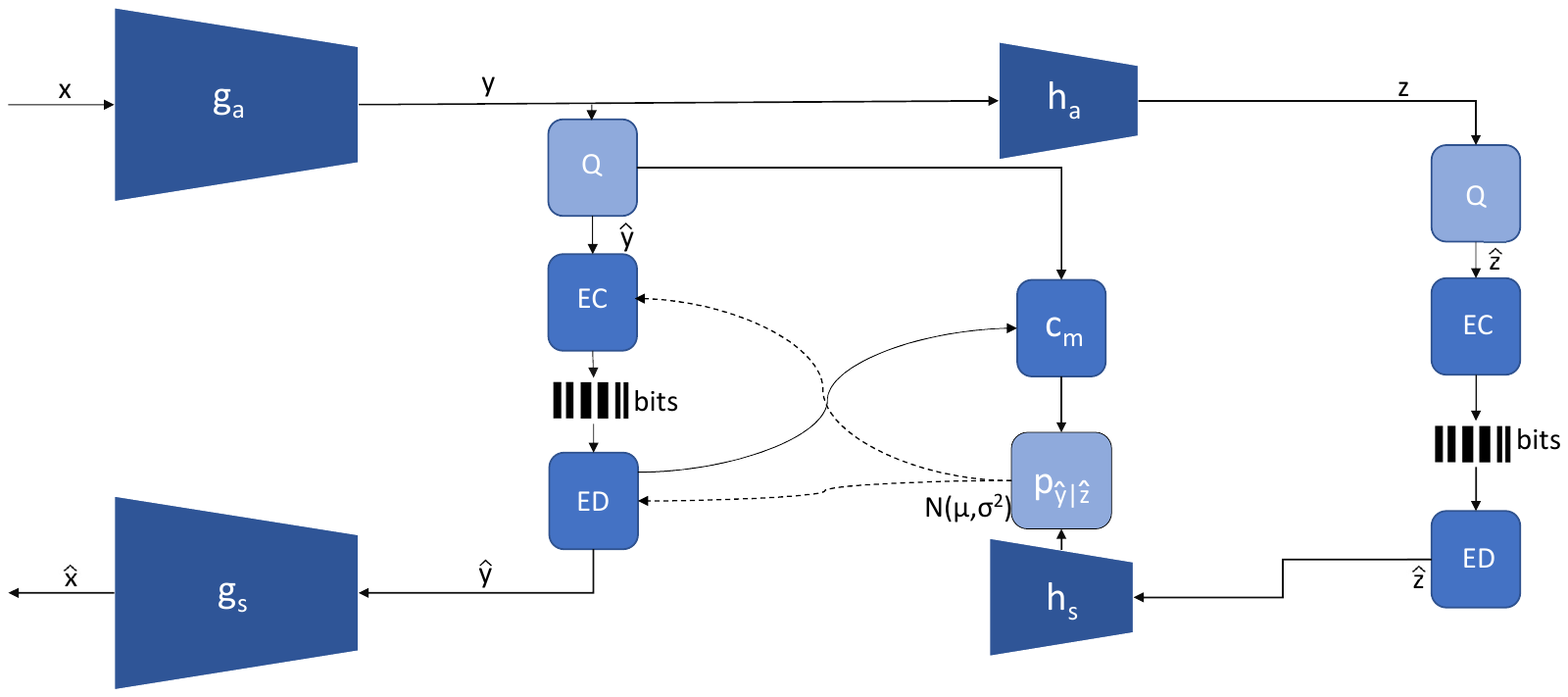}
\caption{\protect\label{fig:usual_architecture}The end-to-end learned image compression architecture of \cite{Minnen2018JointAA}. The analysis and synthesis $g_a$ and $g_s$ handles the transforms between image space and latent space of reduced dimension. Hyperprior analysis $h_a$ and synthesis $h_s$ transform captures the contextual information. The quantization $Q$ and the entropy coding and decoding $EC$ and $ED$ converts the latent vector into a compact binary stream. Context module $c_m$ and probability distribution of latent variables $p_{\hat y|\hat z}$ estimate the distribution of latent variable $\hat y$ conditioned on side information $\hat z$.}
\vspace{-0.5cm}

\end{figure}
Fig. \ref{fig:usual_architecture} provides a high-level overview of a state-of-the-art LIC architecture. For encoding, an analysis transform module \(g_{a}\) transforms the image $x$ into a latent variable $y$ as shown in Equation \ref{eq:1}. Subsequently, $y$ is quantized to produce the discrete representation of the latent variable, $\hat{y}$. Then, $\hat{y}$ is compressed into bitstreams using entropy coding methods such as arithmetic coding \cite{1056282}. We utilize Balle et al.'s method \cite{Ball2016EndtoendOI} to use the quantized latent variables by introducing a uniform noise $U$ (${\text -}0.5, 0.5$) to $y$ during training, where $U$ denotes a uniform distribution centered on $y$. To simplify the process, we denote both the latent features with added uniform noise during training and the discretely quantized latent variables during testing as $\hat{y}$. For decoding, a synthesis transform module \(g_{s}\) reconstructs the quantized variables $\hat{y}$ back to the image $\hat{x}$.

\begin{equation}\label{eq:1}
y = g_a(x),  
\hat{y} = Q(y), 
\hat{x} = g_s(\hat{y})
\end{equation}

The latent variables $\hat{y}$ are modeled as guassian distribution with standard deviation $\sigma$ and mean $\mu$ and then combined with additional side or contextual information $\hat{z}$, as demonstrated in Equation \ref{eq:2}. The distribution of $\hat{y}$ is based on Semi Global Matching (SGM) based entropy model \cite{Minnen2018JointAA}.

\begin{equation} \label{eq:2}
p_{\hat{y}|\hat{z}} (\hat{y}|\hat{z}) = N(\mu, \sigma^2) 
\end{equation}

The main goal of LIC methods is to minimize the weighted sum of the tradeoff between rate and distortion during training:
\begin{equation}
L = R(\hat{y}) + \lambda \ D(x, \hat{x})
\end{equation}

The rate $R$ represents the bit rate of latent variables $\hat{y}$ and $\hat{z}$, which is estimated by the entropy model during training. The distortion $D$ is defined as $D$ = MSE($x$, $\hat{x}$) for MSE optimization and $D$ = 1 ${\text -}$ MS${\text -}$SSIM($x$, $\hat{x}$) for MS${\text -}$SSIM \cite{1292216} optimization. We use $\lambda$ to control the rate-distortion tradeoff across various bit rates. Different $\lambda$ values are discussed in Section \ref{trainingdetails}.
\begin{figure*}
\centering

\includegraphics[width=1\textwidth, trim = 0cm 10cm 0cm 3cm]{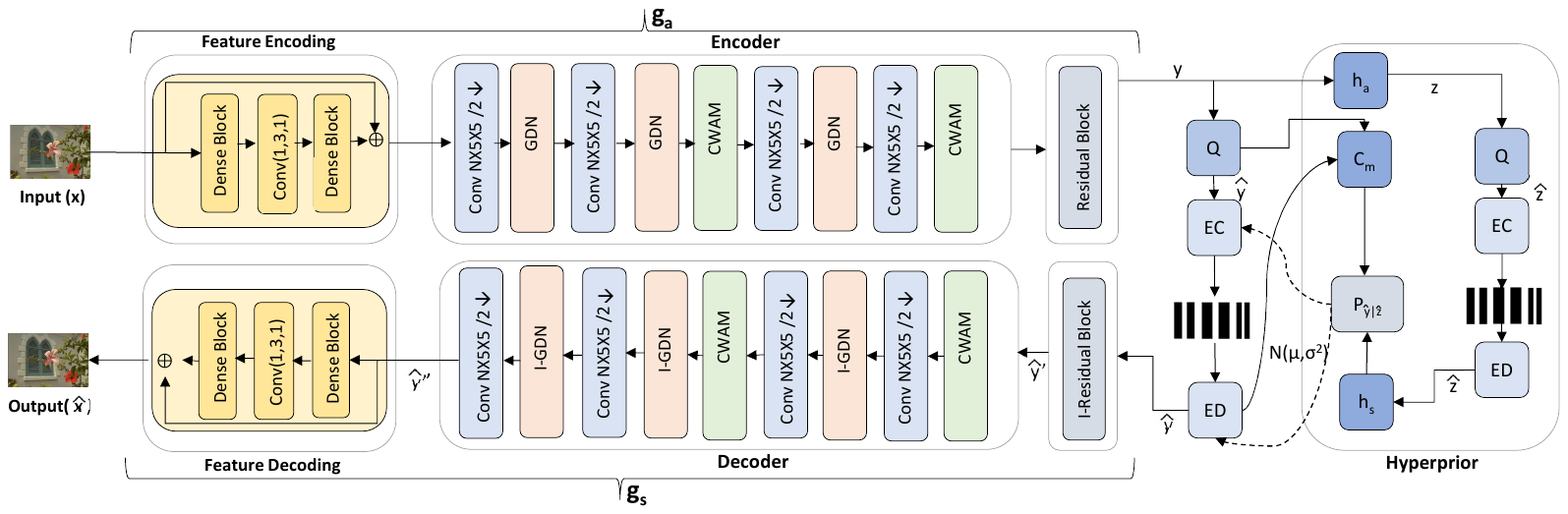}
\caption{\protect\label{fig:architecture} The architecture of our image compression network is based on \cite{9190935}. The analysis transform $g_a$ and synthesis transform $g_s$ convert variables from image space (x) to latent space (y) and from latent space ($\hat{y}$) to image space ($\hat{x}$) respectively. The feature encoding module enhances image features. The encoder and decoder consist of convolutional layers with 5 $\times$ 5 kernel and N channels (set to 320), GDN, and CWAM. IGDN represent the inverse GDN module. EC and ED represent the arithmetic encoder and arithmetic decoder, respectively. $h_a$ and $h_s$ are the hyperprior analysis and synthesis transforms implemented in Minnen et al. \cite{ballé2018variational}. The residual block comprises of 1$\times$1 and 3$\times$3 convolutional layers with CWAM.}
\end{figure*}
\vspace{-0.5in}
\subsection{\textbf{Proposed Method}}
Fig. \ref{fig:architecture} illustrates our network architecture. The proposed method focuses on enhancing the analysis $g_a$ and synthesis $g_s$ transforms between the image space $x$ and the latent feature space $y$. We leverage the existing work for hyper-prior architecture \cite{Minnen2018JointAA,ballé2018variational} and auto-regressive entropy model \cite{Minnen2020ChannelWiseAE} to optimize the distribution of latent features. To enhance the analysis and synthesis modules, we divide the architecture into three sub-modules: Feature Encoding and Feature Decoding, Encoder for image space ($x$ to latent space $y$ conversion), Decoder for latent space ($\hat{y}$ to image space $\hat{x}$ conversion), and Residual Block and I-Residual Block.

  


\vspace{-0.5cm}
\subsubsection{\textbf{Feature Encoding and Feature Decoding}}
While CNNs are powerful at modeling transformations, they struggle to effectively represent the challenging parts. Therefore, we incorporate a residual feature encoding module before the encoder and a feature decoding module after the decoder.The feature encoding module enhances the representativeness of our network by focusing more on the challenging parts of the image and reducing bits for simpler parts. Conversely, the feature decoding module decodes and enhances the reconstruction generated by the decoder. Both modules are built on the popular Dense Block \cite{huang2018densely}, utilizing three cascade convolutions with kernel size 1, 3, 1. 

\vspace{-0.5cm}
\subsubsection{\textbf{Encoder and Decoder}} 
We construct our encoder and decoder modules using four convolutional layers, each with $N$ channels set to 320 and a 5 $\times$ 5 kernel. To enhance these modules, we incorporate GDN and CWAM layers into multiple segments of the network. While the feature encoding module improves the network's representativeness, CWAMs allocate bits more efficiently across different areas internally, albeit with some computational overhead. Despite its simplicity, this architecture can significantly enhance the rate-distortion performance.

\vspace{0.2cm}

\par{\textbf{Cross Window-based Attention Module (CWAM)}} \textcolor{mycolor}{The Cross${\text -}$Window Attention Module (CWAM) \cite{lu2022video} was originally proposed for video interpolation. We introduce this module into LIC architecture to effectively expand the receptive field. In this approach, the input feature map F is downscaled by half to produce a reduced version. This downscaled version is then divided into non-overlapping sub-windows. To facilitate this process, we use the reflection mode padding with specific dimensions before segmenting it into overlapping blocks of a defined size.}



\textcolor{mycolor}{The interaction between windows from the fine-scale feature F and the coarse-scale feature F↓ integrates multi-scale information, resulting in more comprehensive feature representation. Conversely, windows in F↓ cover a larger context compared to those in F. For example, a window Y in F↓ covers four times the context of a corresponding window X in F. This effectively enlarges the receptive field of self-attention.}

The architecture of Residual Block and Inverse Residual Block (I-Residual Block) is similar to that of Cheng et al. \cite{Cheng2020LearnedIC}. It consist of three cascade convolutions with kernel size 1, 3, 1.

\section{\textbf{Experiments and Results}}
\vspace{-0.2cm}
\subsubsection{\textbf{Dataset}} For training, we utilize the Flicker 2W dataset as used in \cite{Liu2020AUE}, which consists of 20,745 high-quality general images. We select approximately 200 images randomly for our validation set, while the remaining images are used for training. Next, we prepare 256 × 256 randomly cropped patches from these images. Finally, we train our network on these patches using the advanced CompressAI PyTorch library \cite{Bgaint2020CompressAIAP}. It is important to note that we exclude a few images with a height or width smaller than 256 pixels for simplicity. For evaluation, we use the commonly used Kodak image dataset \cite{kodak} and CLIC validation dataset \cite{clic}. The Kodak dataset contains 24 uncompressed images with resolutions of 768 × 512, while the CLIC dataset includes 30 high-quality images with much higher resolutions of 1152 × 2048 or higher.
\vspace{-0.4cm}
\subsubsection{\textbf{Training Details}} \label{trainingdetails} All the experiments are conducted on a single Nvidia TITAN X GPU and trained for 600 epochs with a batch size of 4 using Adam optimizer \cite{Kingma2014AdamAM}. Initially, our network is optimized for 450 epochs with an initial learning rate of $10^{\text - 4}$. Subsequently, the learning rate is reduced to $10^{\text - 5}$ at epoch 450 and further decreased to $10^{\text - 6}$ at epoch 550. Our models are optimized using two quality metrics: Mean Squared Error (MSE) and Multi-Scale Structural Similarity Index Measure (MS-SSIM). Following the settings in \cite{Ball2016EndtoendOI}, when optimizing the model for MSE, $\lambda$ is selected from {0.0045, 0.00975, 0.0175, 0.0483, 0.09, 0.14}. When optimizing the model for MS-SSIM, $\lambda$ is chosen from {8.73, 31.73, 60.50}.



\vspace{-0.4cm}
\subsubsection{\textbf{Rate-Distortion Performance}}
\textcolor{mycolor}{Our evaluation involves benchmarking our method against state-of-the-art learned image compression models proposed by Ballé et al. \cite{Ball2018VariationalIC}, Minnen et al. \cite{Minnen2018JointAA}, Lee et al. \cite{Lee2018ContextadaptiveEM}, Hu et al. \cite{Hu2020CoarsetoFineHM}, and Cheng et al. \cite{Cheng2020LearnedIC}. We gather their respective rate-distortion data points from published papers and official GitHub repositories.} Additionally, we compare our approach with \textcolor{mycolor}{widely used traditional} image compression codecs including JPEG \cite{Wallace1991TheJS}, JPEG2000 \cite{Taubman2013JPEG2000I}, WebP \cite{webp}, BPG \cite{bpg}, and VVC \cite{vvc}, assessing their performance using the CompressAI evaluation platform. In case of VVC, we utilize the latest VVC official Test Model VTM 12.1 with an intra-profile configuration sourced from its official GitHub page. \textcolor{mycolor}{We also evaluate BPG using the} BPG software configured with YUV444 subsampling, the HEVC x265 implementation, and an 8-bit depth for image testing. We measure image distortion using peak signal-to-noise ratio (PSNR) and MS-SSIM \cite{Wang2003MultiscaleSS}, and rate performance using bits per pixel (bpp). We generate rate-distortion (RD) curves based on their performance to compare the coding efficiency of different methods.

\begin{figure*}

\setkeys{Gin}{width=.49\linewidth} 
\begin{minipage}[t]{1\columnwidth}
  \subfloat[]{\label{fig:kodak_mse}\includegraphics{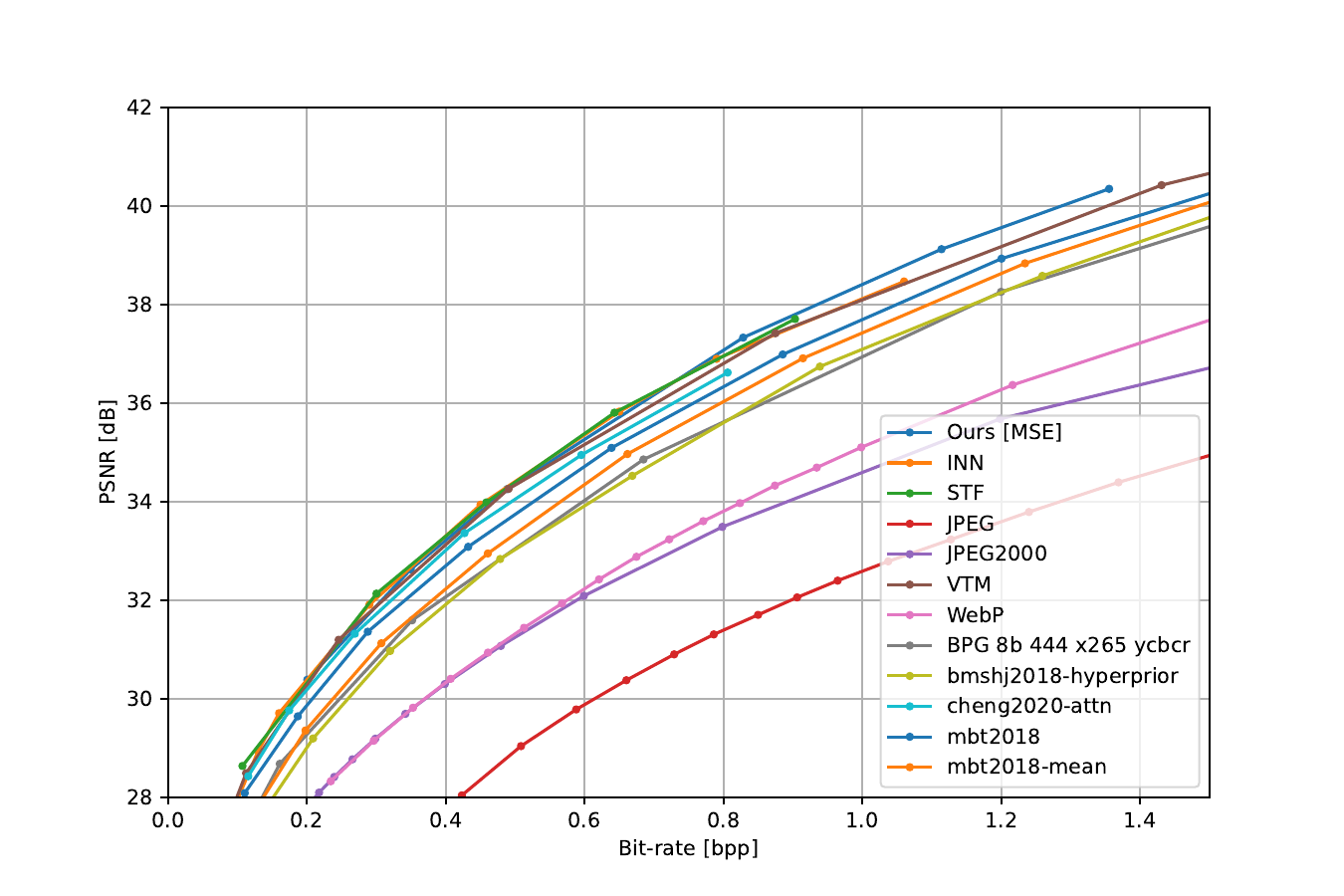}} \hfill
  \subfloat[]{\label{fig:kodak_msssim}\includegraphics{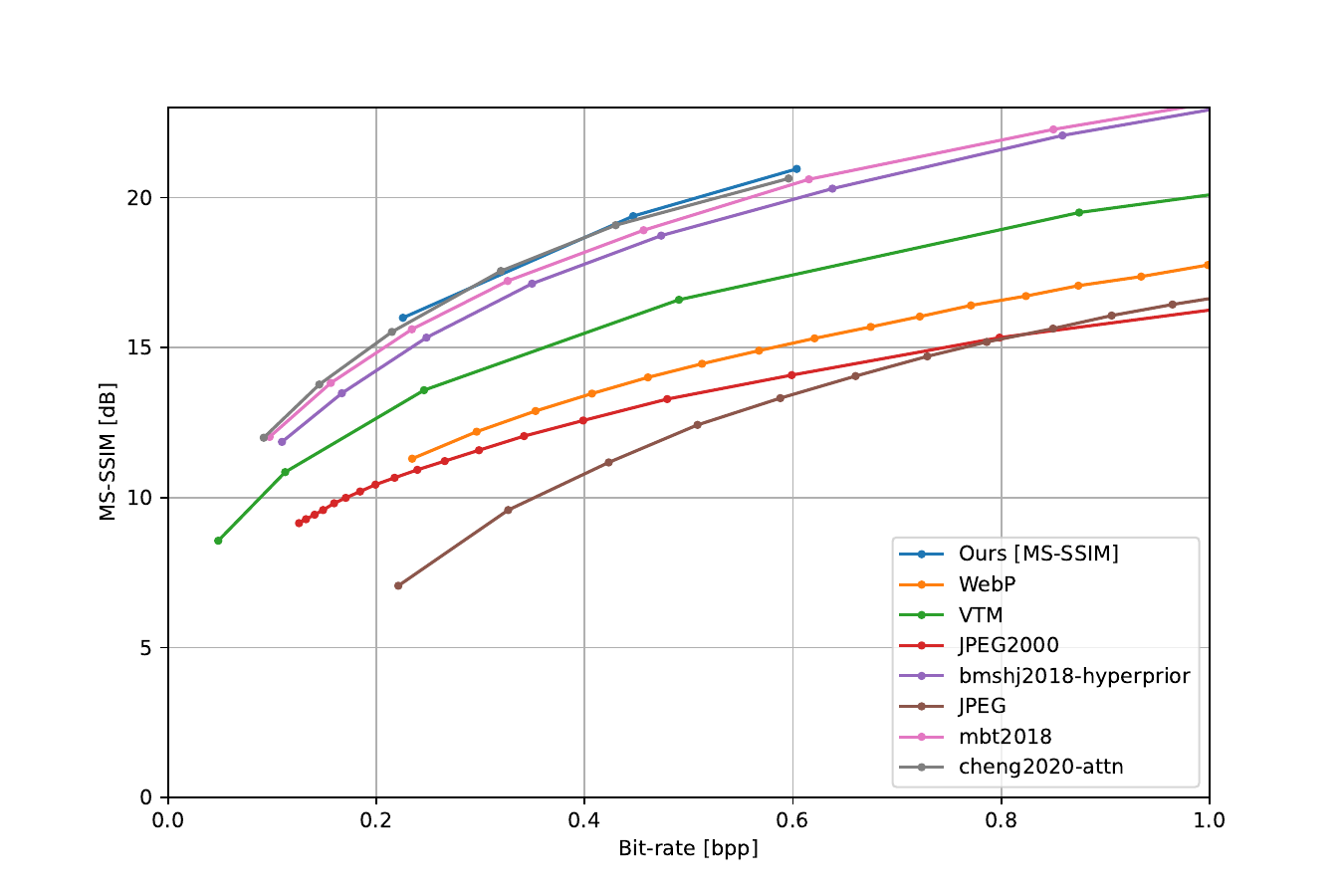}}\hfill
  \vspace{-0.18in}
  \subfloat[]
{\label{fig:clic_mse}\includegraphics{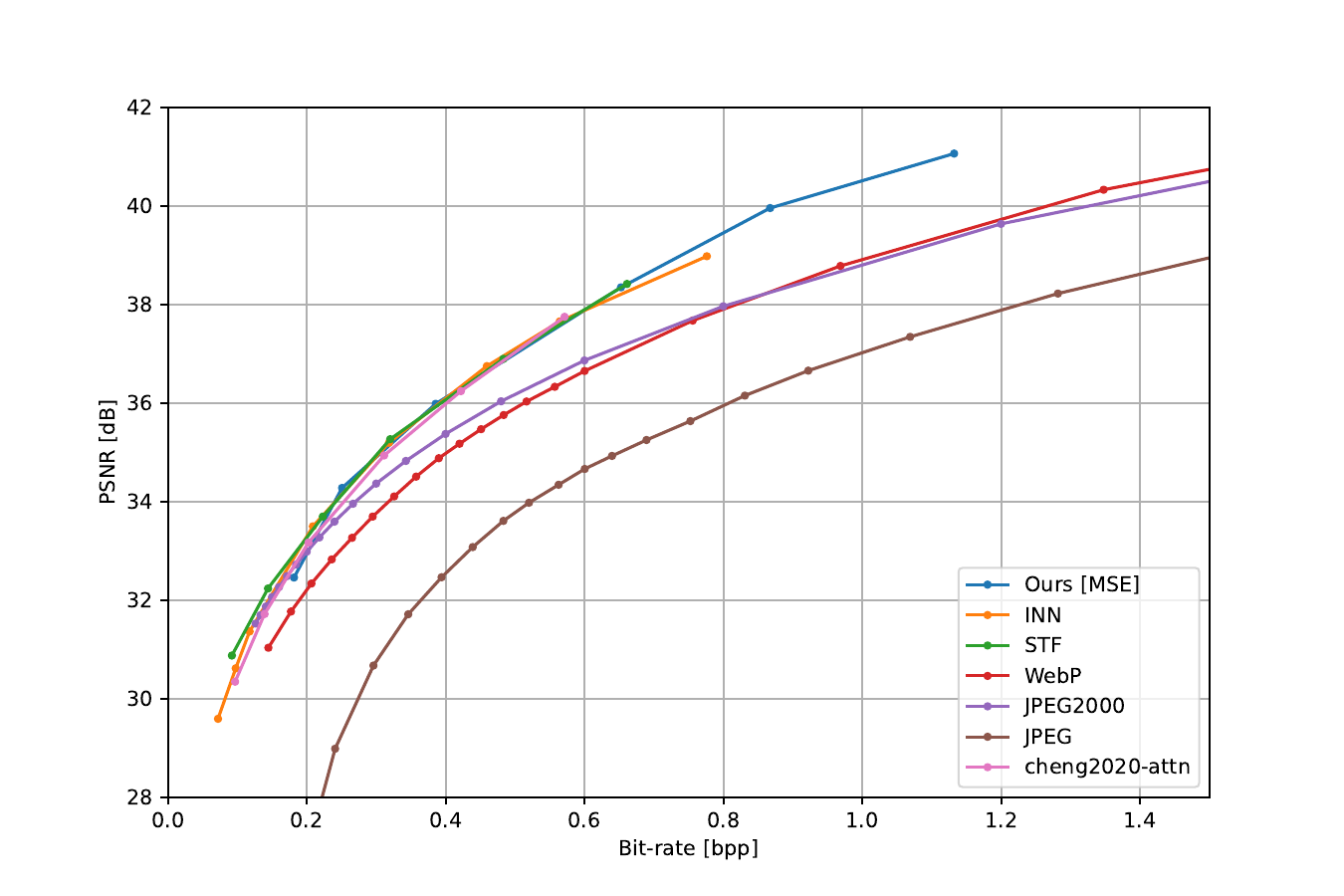}}\hfill
  \subfloat[]{\label{fig:clic_msssim}\includegraphics{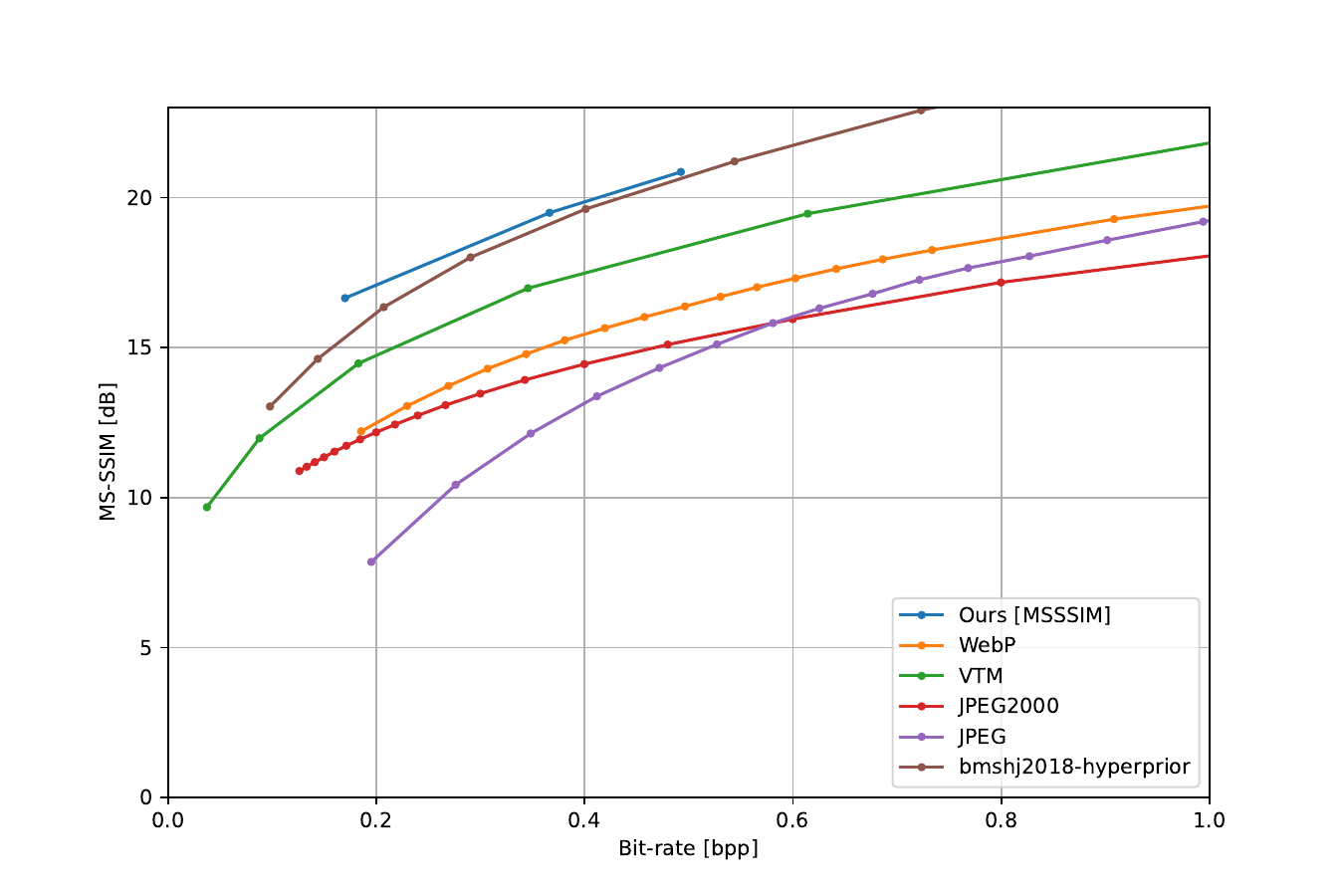}}\hfill
  \caption{RD Performance on Kodak dataset, which contains 24 high quality images \textcolor{mycolor}{(top row)} and on CLIC dataset, which contains 30 high resolution and high quality images \textcolor{mycolor}{(bottom row)}. Our method yields a much better performance when compared with state-of-the-art learned methods and traditional image compression standards. Also note that most images in the CLIC dataset are of high resolution, implying that our method is more robust and promising to compress high-resolution images.} \label{fig:our_results}
\end{minipage}
\end{figure*}

Fig. \ref{fig:our_results} shows the RD curve on the Kodak and CLIC datasets. We convert MS${\text -}$SSIM to ${\text -}$10 log10 (1${\text -}$ MS${\text -}$SSIM) for \textcolor{mycolor}{clarity in comparison}, which is similar to the previous work \cite{Cheng2020LearnedIC}. Our method exhibits a \textcolor{mycolor}{slight edge over} VVC (VTM 12.1) and demonstrates markedly superior performance compared to both established learned methods and traditional image compression standards. For the CLIC dataset, we compare our MSE optimized results with traditional compression standards and the learned methods with official testing results available in their paper or their official GitHub pages. The RD curves on the CLIC dataset are illustrated in Fig. \ref{fig:our_results}. It is evident that our MSE-optimized approach outperforms all other methods. It is worth noting that the majority of images in the CLIC dataset have high resolutions, indicating that our method is more robust for compressing high-resolution images.
\vspace{-0.4in}
\vspace{-1.0cm}
\subsubsection{\textbf{Visual Quality Results}}
\begin{figure}[H]
\centering
\includegraphics[scale=1.5, width=1.1\textwidth, trim = 2.5cm 13cm -0.2cm 2cm]{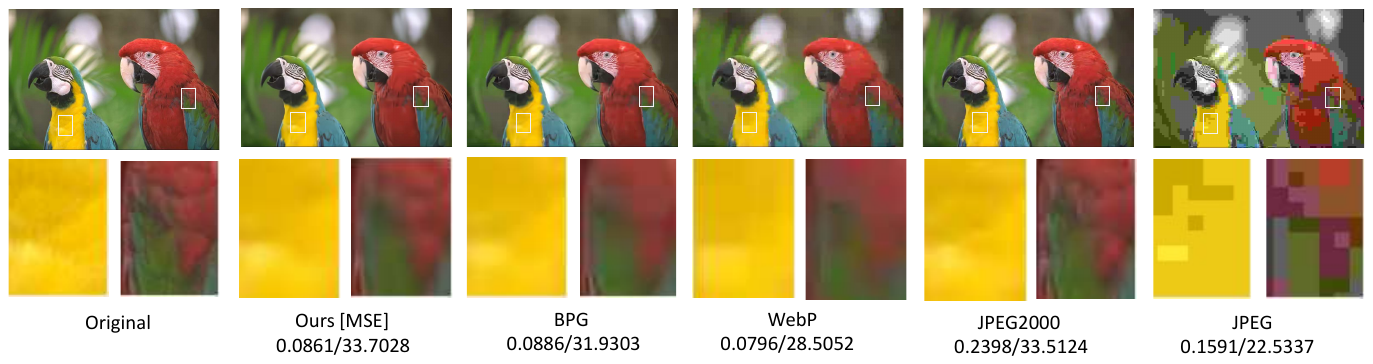}
\caption{\protect\label{fig:ourresults}Reconstructed images from Kodak dataset. The compressed image quality by our method shows better PSNR while maintaining or reducing the BPP in comparison to traditional methods. Subtitles represent BPP$\downarrow$/PSNR$\uparrow$.}
\end{figure}

\begin{figure}[H]
\centering
\includegraphics[scale=1.5, width=1.12\textwidth, trim = 2.5cm 3cm -1.5cm 5cm]{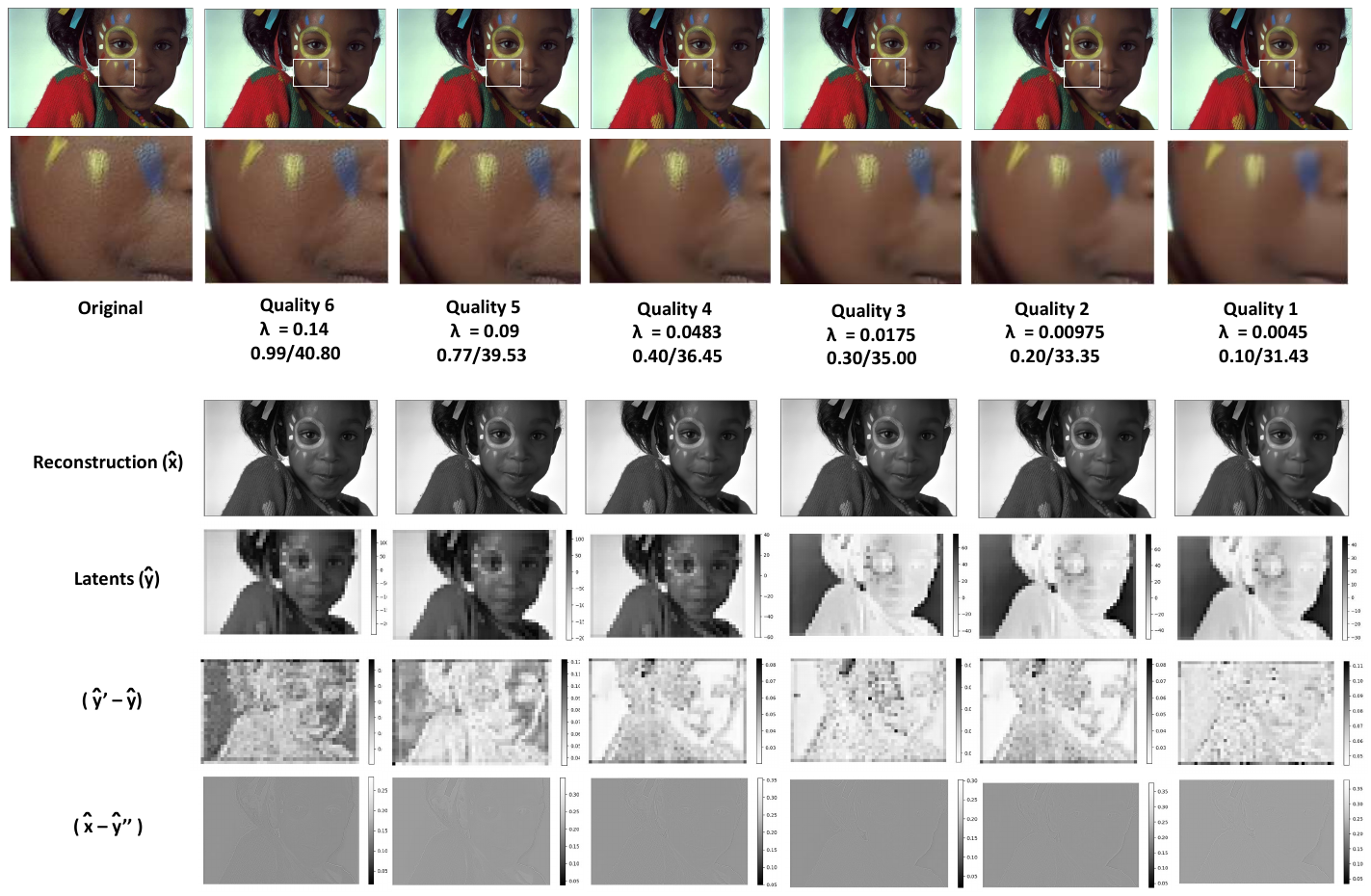}
\caption{\protect\label{fig:ourmethod}Our results at various quality levels of kodim15 from Kodak dataset. Subtitles represent BPP$\downarrow$/PSNR$\uparrow$. Third row represents the visualization of reconstruction in grayscale. Fourth row shows the latents ($\hat{y}$) for channel with maximal entropy. Fifth row represents deviation map between I-residual module ($\hat{y'}$) and latent ($\hat{y}$). Sixth row shows the deviation map between feature decoding module ($\hat{x}$) and decoder ($\hat{y''}$).}
\end{figure}

We present a qualitative comparison of several sample reconstructed images from the Kodak dataset in Fig. \ref{fig:ourresults}.  For JPEG and JPEG2000, we use the lowest quality settings since they cannot achieve the specified bpp levels. Our MSE-optimized method demonstrates commendable performance compared to the latest BPG codec and outperforms other codecs. Additionally, we showcase our results on kodim15 across six different bit rates and qualities in Fig. \ref{fig:ourmethod}. Clearly, images with higher bpp exhibit quality closer to the original image. We depict the deviation map as the difference between latent space variables and variables after the residual block ($\hat{y'}$ - $\hat{y}$) in the fifth row of Fig. \ref{fig:ourmethod} and difference between feature decoding module and decoder output ($\hat{x}$ - $\hat{y''}$) in the sixth row of Fig. \ref{fig:ourmethod}. It is evident that the reconstruction is improved after processing by both modules.

\begin{table}[htp]
\setlength{\tabcolsep}{3pt}
\caption{The complexity of learned image compression models on Kodak dataset.}
\label{tab:1}
\begin{center}
 \scriptsize
 \begin{tabular}{llcccc}\toprule
        

\thead{Methods} & \thead{Encoding \\ Time (s)} & \thead{Decoding \\ Time (s)} & {gMACs} & \thead{Parameters (M)} & \thead{Size (MB)}\\
 \midrule  \addlinespace
Cheng 2020\cite{Cheng2020LearnedIC} & 3.98 & 9.14 & 120.17 & 13.18 & 57\\ 
Hyperprior\cite{Ball2018VariationalIC} & 0.16 & 0.26 & 49.37 & 5.08 & 22 \\
STF\cite{Zou2022TheDI} & 2.346 & 5.212 & 194.99 & 75.24 & 904\\
            INN\cite{Xie2021EnhancedIE} & 2.607 & 5.531 & 272.14 & 50.03 & 209\\
            Ours & 6.291	&  10.298 & 568.62 & 63.17 & 776\\
          
 \bottomrule

\end{tabular}
\label{tab:scoreSnapshot}
\end{center}

\end{table}

\vspace{-0.9cm}

\subsubsection{\textbf{Complexity}}
We assess the complexity and qualitative outcomes of various methods  \cite{Cheng2020LearnedIC,Ball2018VariationalIC,Zou2022TheDI,Xie2021EnhancedIE} using the Kodak dataset as demonstrated in Table \ref{tab:1}. While our method's results suggest that it can outperform these methods in terms of rate-distortion (RD) performance, there is a tradeoff in terms of multiply accumulate-operations (gMACs), size, and encoding/decoding time.




\subsubsection{\textbf{Ablation Study}}
To validate our hypothesis that introducing the Cross-scale
Window-based Attention Module (CWAM) enlarges the receptive field effectively and results in better RD-performance, we conduct several experiments. These experiments involve removing the CWAM and feature encoding or replacing the CWAM with other state-of-the-art (SOTA) methods, namely the Window Attention Module (WAM) proposed by Zou et al. \cite{Zou2022TheDI}.

\begin{figure}
\setkeys{Gin}{ width=1\linewidth} 
\begin{minipage}[t]{1\columnwidth}
  {\label{fig:psnrablationstudy}\includegraphics[trim = 0cm 0.0cm 0cm 0.5cm]{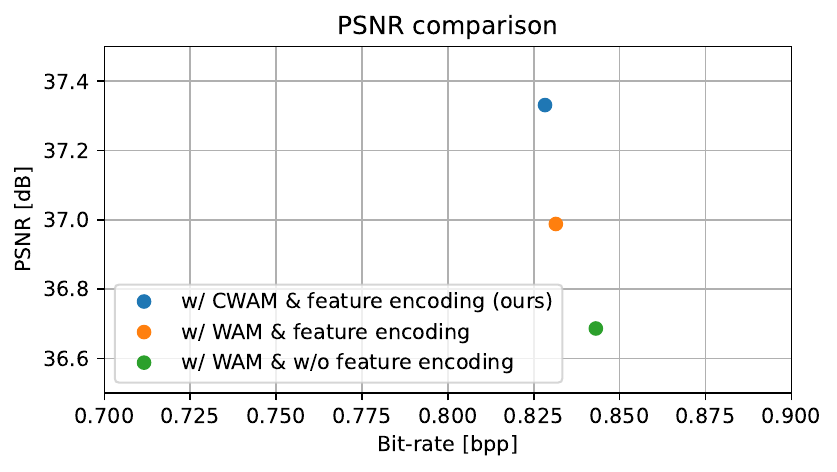}}\hfill

  \caption{Ablation study results. We trained three different models using CWAM with feature encoding, WAM with feature encoding, and WAM without feature encoding for $\lambda$ = 0.0483, when optimized for MSE. Our method with cross window attention and feature encoding module outperforms state-of-the-art window based attention method.} \label{fig:ablation_1}
\end{minipage}\hfill 
\end{figure}

In the first experiment, we replace the proposed CWAM with state-of-the-art method WAM \cite{Zou2022TheDI} and remove the proposed feature encoding module. We train the model for 600 epochs for $\lambda$ = 0.0483. Our results on the Kodak dataset are presented in Fig. \ref{fig:ablation_1}. It is evident that our proposed CWAM, along with the feature encoding module, enhances the model's ability to compress images with low bpp while maintaining higher PSNR.


In the second experiment, we integrate the feature encoding into the network architecture and only replace CWAM with WAM to analyze the contribution of the proposed CWAM. The model is trained for 600 epochs with $\lambda$ = 0.0483. Fig. \ref{fig:ablation_1} illustrates that this experiment yielded lower PSNR with a slightly higher bpp compared to CWAM when analyzed on the Kodak dataset.


\section{\textbf{Discussion}}
While our architecture outperforms state-of-the-art methods in terms of RD performance, there is still room for improvement. The proposed cross-window-based attention is slow due to its attempt to capture a wider receptive field. Our architecture could be optimized in terms of size, GMacs, and parameters, as it is evident that some previous studies outperform ours. One possible reason could be the increased number of channels in our network architecture. Our future experiments should aim to optimize the model size, encoding and decoding time, parameters, and GMacs while continuing to enhance the RD performance.

\section{\textbf{Conclusion}}
\vspace{-0.2cm}

In this paper, we have presented two novel components: a cross window-based attention module to capture correlations among spatially neighboring windows, covering a wider receptive field, and a feature encoding module that captures the representation of challenging portions of an image. Both components are modular and compatible with any architecture for further enhancements. Our extensive experimental results demonstrate the effectiveness of our proposed method, which performs comparably to state-of-the-art methods in terms of rate-distortion performance. This work has the potential for further optimization in terms of improving RD performance, model size, latency, and parameters in the future.
%
%
%
%

\end{document}